\pdfoutput=1
\documentclass[11pt]{article}

\usepackage{acl}

\usepackage{times}
\usepackage{latexsym}

\usepackage[T1]{fontenc}
\usepackage[utf8]{inputenc}

\usepackage{microtype}

\usepackage{inconsolata}

\usepackage{amsmath, amssymb}
\usepackage{amsfonts}  % some math symbol  for example, /mathbb{}
\usepackage{graphicx}  % resize tables and figures
\usepackage{boldline}  % for tables
\usepackage{multirow}  % for table
\usepackage{floatrow}  % resize table for half page.
\usepackage{subcaption} % for figures
\usepackage{makecell}

\newcommand\blfootnote[1]{%
  \begingroup
  \renewcommand\thefootnote{}\footnote{#1}%
  \addtocounter{footnote}{-1}%
  \endgroup
}

\title{Block-Diagonal Orthogonal Relation and Matrix Entity for Knowledge Graph Embedding}

\author{Yihua Zhu$^{1,2}$ \qquad  Hidetoshi Shimodaira$^{1,2}$ \\
  $^1$Kyoto University \qquad $^2$RIKEN\\
  \texttt{zhu.yihua.22h@st.kyoto-u.ac.jp, shimo@i.kyoto-u.ac.jp}
  }

\date{}

\begin{document}
\maketitle

%................................abstract
\begin{abstract}
    The primary aim of Knowledge Graph Embeddings (KGE) is to learn low-dimensional representations of entities and relations for predicting missing facts. Although rotation-based methods like RotatE \cite{sun2019rotate} and QuatE \cite{zhang2019quaternion} perform well in KGE, they face two challenges: limited model flexibility requiring proportional increases in relation size with entity dimension, and difficulties in generalizing the model for higher-dimensional rotations. 
    To address these issues, we introduce OrthogonalE, a novel KGE model employing matrices for entities and block-diagonal orthogonal matrices with Riemannian optimization for relations. This approach not only enhances the generality and flexibility of KGE models but also captures several relation patterns that rotation-based methods can identify. 
    Experimental results indicate that our new KGE model, OrthogonalE, offers generality and flexibility, captures several relation patterns, and significantly outperforms state-of-the-art KGE models while substantially reducing the number of relation parameters.
\end{abstract}

\blfootnote{Our code is available at \url{https://github.com/YihuaZhu111/OrthogonalE}.}

% Introduction
\section{Introduction} \label{sec:intro}

%.....figure in the introduction
\begin{figure}[t]
\centering
\begin{subfigure}[b]{1\textwidth}
        \includegraphics[width=\textwidth]{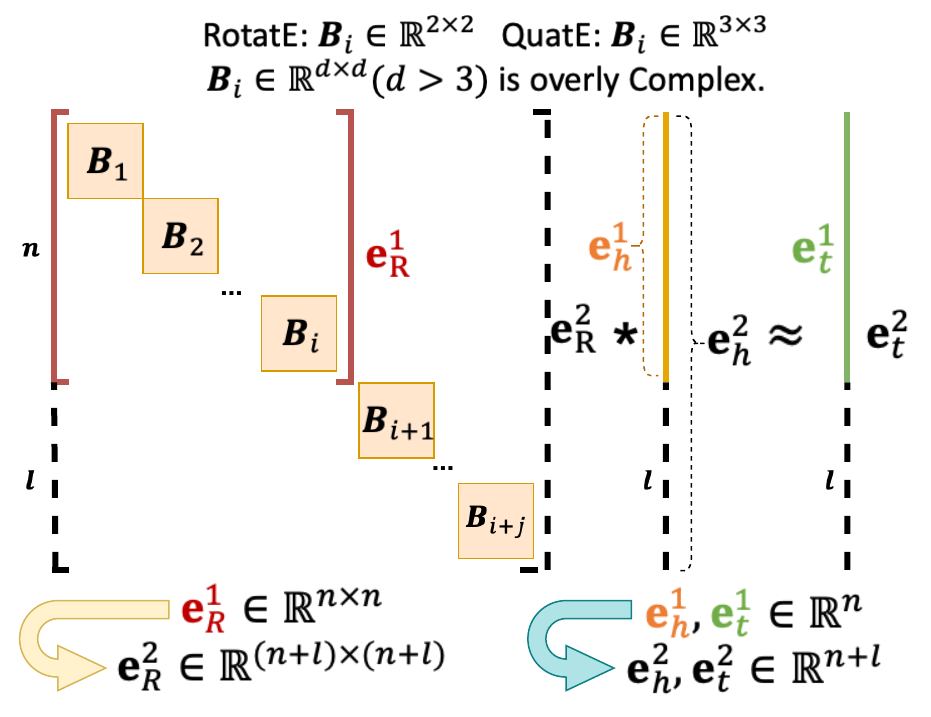}
    \end{subfigure}
\caption{Fundamental operations ($\mathbf{e}_{R}^{1} \cdot \mathbf{e}_{h}^{1} \approx \mathbf{e}_{t}^{1}$) and inherent challenges of rotation-based KGE models. Rotation-based methods require increasing relation parameters for adequate entity representation (lack of flexibility) and struggle with researching higher-dimensional rotation embeddings ($d>3$) due to their complexity (lack of generality). OrthogonalE, depicted in Fig.~\ref{fig:approach_OrthogonalE}, efficiently resolves these challenges.}
\label{fig:problem_RotatE}
\end{figure}

%......figure in the introduction

The fundamental elements of knowledge graphs (KGs) are factual triples, each represented as $(h, r, t)$, indicating a relationship $r$ between head entity $h$ and tail entity $t$. Notable examples include Freebase \cite{bollacker2008freebase}, Yago \cite{suchanek2007yago}, and WordNet \cite{miller1995wordnet}. KGs have practical applications in various fields such as question-answering \cite{hao2017end}, information retrieval \cite{xiong2017explicit}, recommender systems \cite{zhang2016collaborative}, and natural language processing \cite{yang2019leveraging}, garnering considerable interest in academic and commercial research.

Addressing the inherent incompleteness of KGs, link prediction has become a pivotal area of focus. Recent research \cite{bordes2013translating,trouillon2016complex} has extensively leveraged Knowledge Graph Embedding (KGE) techniques, aiming to learn compact, low-dimensional representations of entities and relations. These approaches, marked by scalability and efficiency, have shown proficiency in modeling and deducing KG entities and relations from existing facts.

Recently, rotation-based KGE like RotatE \cite{sun2019rotate} and QuatE \cite{zhang2019quaternion} methods have achieved notable success in the field. 
RotatE uses the Hadamard product to multiply the real and imaginary components of the head entity embedding by the angle-based relation embedding, resulting in a 2D rotation effect within each unit. Each unit consists of two elements representing the real and imaginary components. For example, an entity embedding with 500 dimensions consists of 250 units. The resulting vectors in each unit are then concatenated to form the tail entity embedding. The essence of this operation is the matrix multiplication between a 2D rotation matrix and the entity vector, followed by concatenation as well, as illustrated in Fig.~\ref{fig:problem_RotatE}, multiplying the relation matrix $ \mathbf{e}_{R}^{1} \in \mathbb{R}^{n \times n} $ composed of the block-diagonal Rotation matrix $ \mathbf{B}_{i} \in \mathbb{R}^{d \times d} $ (RotatE: $ \mathbb{R}^{2 \times 2} $) with the head entity vector $ \mathbf{e}_{h}^{1} \in \mathbb{R}^{n} $. Here, we treat the concatenation operation as a block-diagonal arrangement of each 2D rotation matrix.
Similarly, QuatE extends RotatE by using quaternions, which consist of one real and three imaginary components, achieving a 3D rotation effect through the multiplication of a 3D rotation matrix with the entity vector. For QuatE, we replace the 2D rotation matrix $\mathbb{R}^{2 \times 2}$ with a 3D rotation matrix $\mathbb{R}^{3 \times 3}$ as shown in Fig.~\ref{fig:problem_RotatE}.

However, these approaches face two primary issues: lack of flexibility and generality, as illustrated in Fig.~\ref{fig:problem_RotatE}. First, the model's lack of flexibility necessitates increasing the size of the overall relation matrix ($ \mathbf{e}_{R}^{1} \in \mathbb{R}^{n \times n} \rightarrow \mathbf{e}_{R}^{2} \in \mathbb{R}^{(n+l) \times (n+l)} $) to meet entity dimension requirements ($ \mathbf{e}_{h}^{1} \in \mathbb{R}^{n} \rightarrow \mathbf{e}_{h}^{2} \in \mathbb{R}^{n+l} $) for better entity representation. For example, when the entity vector changes ($ \mathbf{e}_{h}^{1} \in \mathbb{R}^{100} \rightarrow \mathbf{e}_{h}^{2} \in \mathbb{R}^{1000} $) to improve representation, the parameter increase is 900. However, the corresponding change in the relation matrix ($ \mathbf{e}_{R}^{1} \in \mathbb{R}^{100 \times 100} \rightarrow \mathbf{e}_{R}^{2} \in \mathbb{R}^{1000 \times 1000} $) results in a parameter increase of 990,000. This substantial increase leads to redundancy and inefficiency in representing relations.

Second, the model's lack of generality makes it challenging to explore high-dimensional rotational KGE models due to the significant computational demands and the complexity of rotations in higher dimensions ($ \mathbf{B}_{i}\in \mathbb{R}^{2 \times 2}, \mathbb{R}^{3 \times 3} \rightarrow \mathbb{R}^{d \times d}, d>3 $), such as SO(4), SO(5), and SO(10)\footnote{$\textrm{SO}(d)$ is the set of orthogonal matrices with determinant 1 that represent rotation transformations in $d$-dimensional Euclidean space.}. This restricts the development of more generalized and higher-dimensional rotation KGE approaches.

%.....figure of approach
\begin{figure}[t]
\centering
\begin{subfigure}[b]{1\textwidth}
        \includegraphics[width=\textwidth]{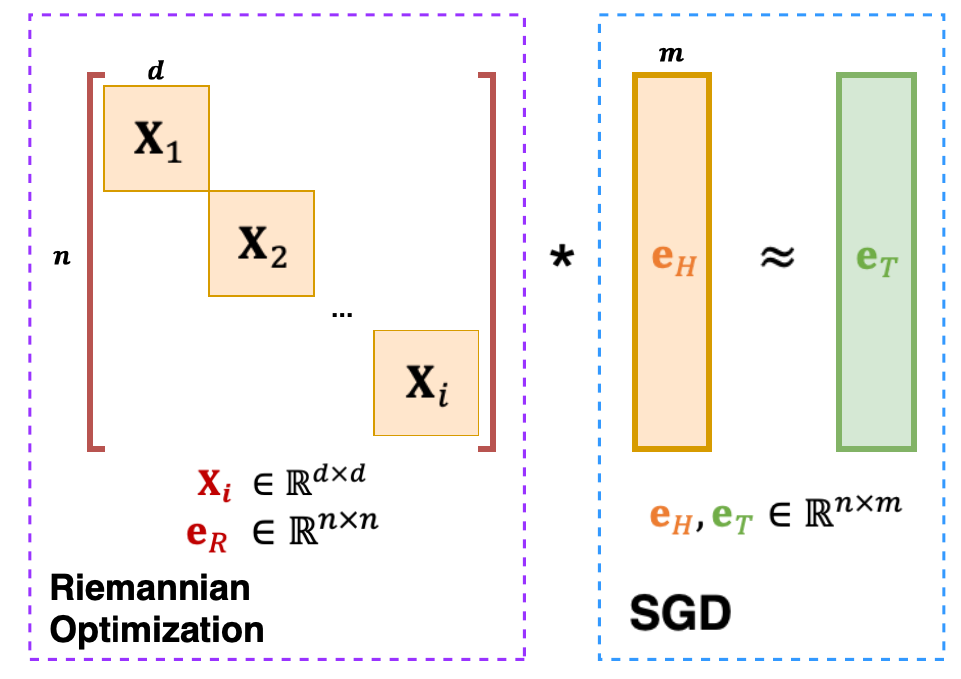}
    \end{subfigure}
\caption{Diagram of the OrthogonalE approach. We employ matrices for entities and block-diagonal orthogonal matrices with Riemannian optimization for relations, thereby retaining the advantages of rotation-based method relation patterns while addressing its two main issues.}
\label{fig:approach_OrthogonalE}
\end{figure}

%......figure of approach

%.....................................table notation
\begin{table}[t]
\resizebox{\textwidth}{!}{\renewcommand{\arraystretch}{1.1}
   \centering
   \begin{tabular}{lc}
   \clineB{1-2}{2}
   {Notation} & {Explanation} \\
   \clineB{1-2}{2}
   $(h, r, t) \in \mathcal{E}$ & Fact triples \\
   $\mathcal{V}$ & Entity sets \\
   $\mathcal{R}$ & Relation sets \\
   $\mathbf{e}_{v} \in \mathbb{R}^{n}$ & Entity vector rep  \\
   $\mathbf{e}_{V} \in \mathbb{R}^{n \times m}$ & Entity matrix rep in OrthogonalE \\
   $\mathbf{e}_{R} \in \mathbb{R}^{n \times n}$ & Relation matrix rep  \\
   $\mathbf{B}_{i} \in \mathbb{R}^{d \times d}$ & Block-diagonal rotation matrix  \\
   $\mathbf{X}_{i} \in \mathbb{R}^{d \times d}$ & Block-diagonal orthogonal matrix  \\
   $n \in \mathbb{R}^{1}$ & Row size of relation matrix rep \\
   $m \in \mathbb{R}^{1}$ & Column size of entity matrix rep \\
   $d \in \mathbb{R}^{1}$ & size of Block-diagonal matrix \\
   $d^{E}\left( ., . \right)$ & Euclidean distance\\
   $b_{v} \in \mathbb{R}^{1}$ & Entity bias\\
   $\cdot$ & Matrix multiplication \\
   $s(h,r,t)$ & Scoring function \\
   \clineB{1-2}{2}
   \end{tabular}
   }
\caption{Notation summary. Within the table, $\mathbf{e}_{v}$ includes the head $\mathbf{e}_{h}$ and tail $\mathbf{e}_{t}$ entity vectors as used in traditional KGE methods, whereas $\mathbf{e}_{V}$ consists of the head $\mathbf{e}_{H}$ and tail $\mathbf{e}_{T}$ entity matrix representations in our OrthogonalE approach. Furthermore, `rep' in the table denotes representation.}
\label{tab:notation}
\end{table}
%....................................table notation

To overcome these two issues, we propose a highly general and flexible KGE model named OrthogonalE as shown in Fig.~\ref{fig:approach_OrthogonalE}, and detailed notation is shown in Table~\ref{tab:notation}. 
Firstly, by transforming entity vectors $ \mathbf{e}_{v} \in \mathbb{R}^{n} $ into matrices $ \mathbf{e}_{V} \in \mathbb{R}^{n \times m} $ to better represent entities, we control the entity dimension through variable $ m $, avoiding unnecessary expansion of the size of the relation matrix. Corresponding to the above example of lack of flexibility, we can maintain the relation matrix size ($ \mathbf{e}_{R} \in \mathbb{R}^{100 \times 100} $) and only modify the entity matrix size ($ \mathbf{e}_{V}^{1} \in \mathbb{R}^{100 \times 1} \rightarrow \mathbf{e}_{V}^{2} \in \mathbb{R}^{100 \times 10}, m=1 \rightarrow 10 $) to meet the requirements of entity representation.
Secondly, leveraging the concept that rotation matrices are orthogonal, we replace rotation matrices $ \mathbf{B}_{i} $ with orthogonal matrices\footnote{$\textrm{O}(d)$ is the set of square orthogonal matrices in $d$-dimensional Euclidean space, including those with a determinant of 1 or -1, representing transformations such as rotations and reflections.} $ \mathbf{X}_{i} \in \mathbb{R}^{d \times d} $ of adaptable dimensions $ d $, facilitating the exploration of higher-dimensional block-diagonal orthogonal matrix models. Lastly, for effective optimization, we employ Riemannian optimization for the relation matrix $ \mathbf{e}_{R} \in \mathbb{R}^{n \times n} $ and Stochastic Gradient Descent (SGD) for the entity matrix  $ \mathbf{e}_{V} \in \mathbb{R}^{n \times m} $.

We evaluate the new model on two KGE datasets including WN18RR \cite{dettmers2018convolutional}, FB15K-237 \cite{toutanova2015observed}. Experimental results indicate that our new KGE model, OrthogonalE, offers generality and flexibility, captures several relation patterns, and significantly outperforms state-of-the-art KGE models while substantially reducing the number of relation parameters.

\section{Related Work} \label{sec:related_work}

\paragraph{Knowledge Graph Embedding}
Translation-based approaches are prominent in KGE, notably TransE \cite{bordes2013translating}, which interprets relations as vector translations.
TransH \cite{wang2014knowledge}, TransR \cite{lin2015learning}, and TransD \cite{ji2015knowledge} represent extensions of the translation-based method, building upon the foundational approach of TransE.
ComplEx\cite{trouillon2016complex} advances this by embedding KGs in a complex space and using the Hermitian product for modeling antisymmetric patterns.
Inspired by ComplEx, RotatE \cite{sun2019rotate} then innovated by treating relations as rotations in a complex vector space, capable of capturing varied relation patterns like \textit{Symmetry}, \textit{Antisymmetry}, \textit{Inversion}, and \textit{Commutative Composition}. 
Following this, QuatE \cite{zhang2019quaternion} employed quaternion operations (3D rotations) for even better expressiveness than RotatE. 
DensE \cite{lu2022dense} employed various techniques for 3D rotation implementation and proposed that 3D rotation could handle the relation pattern of \textit{non-commutative composition}.
HopfE \cite{bastos2021hopfe} seeks to employ SO(4) rather than SO(3) for KG representation, which is directly connected to the generality issue discussed in our research. We are also keen on investigating rotations in higher dimensions. Nonetheless, progressing to SO(5) or even SO(10) poses substantial difficulties.

In addition, our research builds upon existing work in the field. For instance, OTE \cite{tang2019orthogonal} uses a compact block-diagonal orthogonal matrix, similar to our approach, to maintain RotatE's relation patterns and complex relations. Yet, our OrthogonalE model improves stability and performance with Riemannian Optimization, outperforming OTE's Gram-Schmidt process.
GoldE's \cite{li2024generalizing} universal orthogonal parameterization, derived from Householder reflections, offers theoretical guarantees for dimensionality and geometry. However, our focus is on enhancing the generality and flexibility of the KGE models, not just on capturing knowledge graph patterns and heterogeneity.

In conclusion, considering the two major disadvantages of rotation-based methods mentioned in Section~\ref{sec:intro}, we need to refine our model to make it more general and flexible.

\paragraph{Optimization on the orthogonal manifold}
In optimization on the orthogonal manifold, transitioning from $ X^t $ to $ X^{t+1} $ while remaining on the manifold necessitates a method known as retraction \cite{absil2012projection}. Prior research has effectively adapted several standard Euclidean function minimization algorithms to Riemannian manifolds. Notable examples include gradient descent \cite{absil2008optimization,zhang2016first}, second-order quasi-Newton methods \cite{absil2007trust,qi2010riemannian}, and stochastic approaches \cite{bonnabel2013stochastic}, crucial in deep neural network training.

Meanwhile, Riemannian optimization is often used for the orthogonal manifolds, and has made significant progress in deep learning, especially in CNNs and RNNs. \citet{cho2017riemannian} innovatively substituted CNN's Batch Normalization layers with Riemannian optimization on the Grassmann manifold for parameter normalization. Additionally, significant strides in stabilizing RNN training have been made by \cite{vorontsov2017orthogonality,wisdom2016full,lezcano2019cheap,helfrich2018orthogonal}, through the application of Riemannian optimization to unitary matrices.

As this paper primarily focuses on KGE, we do not delve deeply into Riemannian optimization. Instead, we utilize retraction with the exponential map for iterative optimization, as implemented in the Geoopt library \cite{kochurov2020geoopt}.

\section{Problem Formulation and Background} \label{background}

Before presenting our approach, we introduce the KGE problem and provide an overview of optimization on orthogonal manifolds.

\subsection{Knowledge Graph Embedding}

In a KG consisting of fact triples $(h, r, t) \in \mathcal{E} \subseteq \mathcal{V} \times \mathcal{R} \times \mathcal{V}$, with $\mathcal{V}$ and $\mathcal{R}$ denoting entity and relation sets, the objective of KGE is to map entities $v \in \mathcal{V}$ to $k_{\mathcal{V}}$-dimensional embeddings $\mathbf{e}_{v}$, and relations $r \in \mathcal{R}$ to $k_{\mathcal{R}}$-dimensional embeddings $\mathbf{e}_{r}$.

A scoring function $s: \mathcal{V} \times \mathcal{R} \times \mathcal{V} \rightarrow \mathbb{R}$ evaluates the difference between transformed and target entities, quantified as a Euclidean distance:
\[
d^{E}\left( \mathbf{x}, \mathbf{y} \right) = \left\| \mathbf{x}-\mathbf{y} \right\|.
\]

\subsection{Optimization on the orthogonal manifold }

\begin{figure}[t]
\centering
\begin{subfigure}[b]{1\textwidth}
        \includegraphics[width=\textwidth]{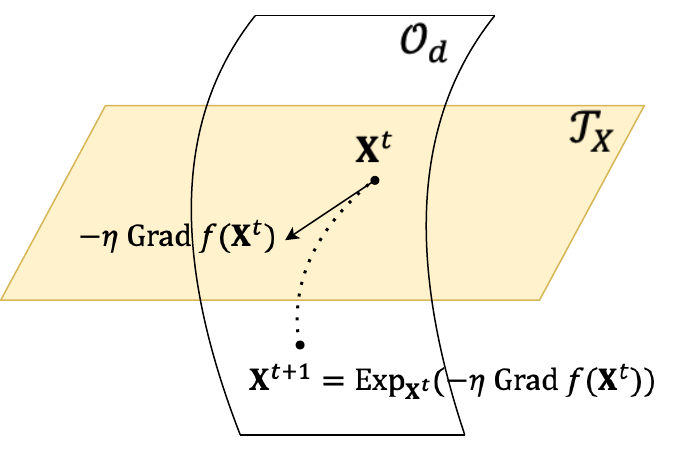}
    \end{subfigure}
\caption{Illustration of Riemannian gradient descent iteration on an orthogonal manifold.}
\label{fig:exponential_map}
\end{figure}

In optimization on the orthogonal manifold, the core problem is formulated as:
\begin{align}
    \min_{X \in \mathcal{O}_d} f(X)   \label{eq:optimization_problem}.
\end{align}
Here, $f$ is a differentiable function mapping elements of $ \mathbb{R}^{d \times d} $ to $\mathbb{R}$, and the \textit{orthogonal manifold} $\mathcal{O}_d$, also denoted as $O(d)$, is defined as $ \mathcal{O}_d \triangleq \left\{ X \in \mathbb{R}^{d \times d} \mid XX^\top = I_{d} \right\}$. 
Moreover, the tangent space at $X$, denoted by $\mathcal{T}_X$, is the set $ \mathcal{T}_X = \left\{\xi \in \mathbb{R}^{d \times d} \mid \xi X^\top + X \xi^\top = 0\right\} $.

To address the problem formulated in~(\ref{eq:optimization_problem}) more efficiently, recent studies suggest optimization of the orthogonal manifold with retractions as an effective approach \cite{ablin2022fast}. In this work, we primarily employ the retraction with exponential map for iterative optimization, as illustrated in Fig.~\ref{fig:exponential_map}. The key iteration formula for this method is:
\begin{align}
    X^{t+1}=\text{Exp}_{X^{t}}(-\eta \,\text{Grad} f(X^{t})),      \label{eq:exponential_iteration}
\end{align}
Where $ t $ indexes the iteration steps, $ \text{Exp}_{X^t}(\xi) $ denotes the exponential map, and $ \eta $ represents the learning rate. $ \text{Grad} f(\cdot) $ is the Riemannian gradient. Subsequent sections will delve into the computation of $ \text{Exp}_{X^t}(\xi) $ and $ \text{Grad} f(\cdot) $.

The exponential map allows movement in a specified direction on the manifold. Starting from $X$ with initial velocity $\xi$, the exponential map on the manifold of orthogonal matrices is given by \cite{massart2022coordinate}:
\[
    \text{Exp}_{X}(\xi) = X \,  \text{expm}(X^\top \xi), \, \forall \xi \in \mathcal{T}_X,   
\]
where $ \text{expm}(\cdot) $ denotes the matrix exponential.

On the orthogonal manifold, the Riemannian gradient $ \text{Grad} f(\cdot) $ is calculated as \cite{absil2008optimization}:
\[
    \text{Grad} f(X) = P_{\mathcal{T}_X}(\nabla f(X)),
\]
where $\nabla f(X)$ is Euclidean gradient of $f(X)$, and the calculation formula for $ P_{\mathcal{T}_X}(\cdot) $ is:
\[
    P_{\mathcal{T}_X}(Y) = X(\frac{X^\top Y - Y^\top X}{2}), Y \in \mathbb{R}^{d \times d}.
\]

\section{Approach}

Our approach is developed  to acquire both a flexible and general KGE model and ensure that this model can concurrently represent several relation patterns.
This is achieved by employing matrices for entities and block-diagonal orthogonal matrices with Riemannian optimization for relations. Figure~\ref{fig:approach_OrthogonalE} illustrates the OrthogonalE approach, and Table~\ref{tab:notation} provides the details of the notations used.

\subsection{Orthogonal Matrices for Relations}

%.............parameter count

\begin{table}[t]
\resizebox{\textwidth}{!}{\renewcommand{\arraystretch}{1.1}
\centering
\begin{tabular}{lcccc}
\clineB{1-5}{2}
\multirow{3}{*}{\centering $\mathbf{e}$} & \multirow{3}{*}{\centering Model} & \multicolumn{3}{c}{Number of Parameters} \\
\clineB{3-5}{1}
 & & {\makecell{Normal \\ }} & {\makecell{$m=1$ \\ \ref{subsec:Orthogonal_Relation} }} & {\makecell{Fixed $\mathbf{e}_{V}$\\ \ref{subsec:Entity}}} \\
\clineB{1-5}{2}
\multirow{3}{*}{{$\mathbf{e}_{V}$}} 
& RotatE & $n$ & $n$ & $n\times m$ \\
& QuatE & $n$ & $n$ & $n\times m$ \\
& OrthogonalE($d\times d$) & $n\times m$ & $n$ & $n\times m$\\
\clineB{1-5}{2}
\multirow{3}{*}{{$\mathbf{e}_{R}$}}
& RotatE & $\frac{n}{2}$ & $\frac{n}{2}$ & $(\frac{n}{2})\times m$ \\
& QuatE & $n$ & $n$ & $n\times m$ \\
& OrthogonalE($d\times d$) & $\frac{(d-1)n}{2}$ & $\frac{(d-1)n}{2}$ & $\frac{(d-1)n}{2}$  \\
\clineB{1-5}{2}
\end{tabular}
}

\caption{The parameter calculations for the KGE models. For all models, the relation matrix size is $n$. The block-diagonal matrix size is 2 for RotatE, 3 for QuatE, and $d$ for OrthogonalE, with an entity matrix column size of $m$ for OrthogonalE. In the table, ``Normal'' represents the standard parameter calculation, ``$m=1$'' constrains the column size of the entity matrix to 1 to explore the impact of block-diagonal orthogonal matrices on the model, as analyzed in section~\ref{subsec:Orthogonal_Relation}. ``Fixed $\mathbf{e}_{V}$'' ensures that the entity dimensions are consistent across all models to demonstrate the parameter savings in the relation matrix when using the entity matrix in OrthogonalE, as discussed in section~\ref{subsec:Entity}.} 
 \label{tab:parameter_count}
\end{table}  
%.............parameter count

To address the challenge of generality for exploring high-dimensional rotational KGE models mentioned in Section~\ref{sec:intro}, we exploit the orthogonality of rotation matrices, substituting rotation matrices ($ \mathbf{B}_{i} \in \mathbb{R}^{d \times d} $) with orthogonal matrices ($ \mathbf{X}_{i} \in \mathbb{R}^{d \times d} $) of corresponding dimensions $d$. Consequently, our relation embedding ($ \mathbf{e}_{R} \in \mathbb{R}^{n \times n} $)  is composed of $ n/d $ block-diagonal orthogonal matrices $ \mathbf{X}_{i} $ as illustrated in Fig.~\ref{fig:approach_OrthogonalE}:
\begin{align}
    \mathbf{e}_{R} = \textrm{diag}(\mathbf{X}_{1},\mathbf{X}_{2},\ldots,\mathbf{X}_{n/d}), \label{eq:relation_embedding}
\end{align}
where the number of relation parameters is
 $\tfrac{d(d-1)}{2} \times \tfrac{n}{d} = \tfrac{(d-1)n}{2} $ as shown in Table~\ref{tab:parameter_count}. This structure allows OrthogonalE to achieve generality, adapting to datasets with diverse complexities by adjusting the dimension $d$ of the block-diagonal matrices. 
Additionally, the employed relation structure facilitates the model's capability to concurrently capture \textit{Symmetry}, \textit{Antisymmetry}, \textit{Inversion}, and \textit{Non-commutative Composition} relation patterns, as substantiated in Appendix~\ref{sec:proof_relation_pattern}, and detailed introduction of relation patterns refer to Appendix~\ref{subsec:relation_pattern_explanations}.

\subsection{Matrix Representation for Entities}
Inspired by \cite{miyato2022unsupervised}, transforms vector embeddings into matrix embeddings to improve embedding effectiveness. In our work, to enhance OrthogonalE's flexibility, we aim to regulate entity dimension using variable $m$ and transform entity vectors $ \mathbf{e}_{v} \in \mathbb{R}^{n} $ into matrices $ \mathbf{e}_{V} \in \mathbb{R}^{n \times m} $ as shown in Fig.~\ref{fig:approach_OrthogonalE}, thus preventing unnecessary expansion of the relation size as shown in Fig.~\ref{fig:problem_RotatE}. This part allows OrthogonalE to acquire flexibility, adapting to diverse datasets with varying relation and entity parameters, rather than indiscriminately increase both. And the number of entity parameters is $ n\times m $.

\subsection{Scoring function and Loss}
We utilize the Euclidean distance between the transformed head entity $ \mathbf{e}_{R} \cdot \mathbf{e}_{H} $ and the tail entity $ \mathbf{e}_{T} $ as the scoring function:
\begin{align}
    s(h,r,t) = -d^{E}\left( \mathbf{e}_{R} \cdot \mathbf{e}_{H}, \mathbf{e}_{T} \right) + \! b_{h} \! + \!b_{t}. 
\end{align}
Here, $b_{v}\,(v \in \mathcal{V})$ denotes the entity bias, incorporated as a margin in the scoring function, following methodologies from \citet{tifrea2018poincar, balazevic2019multi}. Furthermore, we opt for a uniform selection of negative samples for a given triple $(h, r, t)$ by altering the tail entity, rather than employing alternative negative sampling techniques.
The loss function defined as follows:
\begin{align}
L =\sum_{t^{\prime}} \log \left(1+\exp \left(y_{t^{\prime}} \cdot s\left(h, r, t^{\prime}\right)\right)\right) \label{eq:loss-function} \\
y_{t^{\prime}}=\left\{\begin{array}{l}
-1, \text { if } t^{\prime}=t  \nonumber \\
1, \text { otherwise},
\end{array}\right.
\end{align}
where $t^{\prime}$ represents sampled tail entities that include both positive and negative samples.

\begin{table*}[t]
\resizebox{0.9\textwidth}{!}{\renewcommand{\arraystretch}{1}
    \centering
    \begin{tabular}{lcccccccc}
    \clineB{1-9}{2}
    & & \multicolumn{2}{c}{WN18RR} & \multicolumn{5}{c}{FB15K-237}  \\
    
     Model  & MRR & H@1 & H@3 & H@10  & MRR & H@1 & H@3 & H@10\\
     \clineB{1-9}{2}
        TransE $\diamondsuit$   & .226 & - & - & .501 &  .294 & - & - & .465 \\ 
        DistMult $\diamondsuit$ & .430 & .390 & .440 & .490 & .241 & .155 & .263 & .419\\
        ComplEx $\diamondsuit$  & .440 & .410 & .460 & .510 & .247 & .158 & .275 & .428\\
        ConvE $\diamondsuit$   & .430 & .400 & .440 & .520 & .325 & .237 & .356 & .501 \\
        RotatE $\diamondsuit$  &  .470 & .422 & .488 & .565 &  .297 & .205 & .328 & .480 \\ 
        QuatE $\diamondsuit$   &  .481 & .436 & .500 & .564 &  .311 & .221 & .342 & .495 \\ 
        HopfE \cite{bastos2021hopfe} &  .472 & .413 & .500 & \textbf{.586} & \textbf{.343} & \textbf{.247} & \textbf{.379} & \textbf{.534} \\
        DensE \cite{lu2022dense} &  .486 &   -  &  -   & .572 & .306 &  -   &   -  & .481 \\
    \clineB{1-9}{1}
        Gram-Schmidt(2$\times$2) & .475 & .434 & .489 & .556 & .317 & .226 & .344 & .502 \\
        Gram-Schmidt(3$\times$3) & .487 & .445 & .500 & .568 & .322 & .232 & .350 & .504 \\
        \textbf{OrthogonalE}(2$\times$2)& .490 & .445 & .503 & .573 & .330 & .239 & \underline{.368} & .516 \\
        \textbf{OrthogonalE}(3$\times$3)& \underline{.493} & \textbf{.450} & \textbf{.508} & \underline{.580} & .331 & .240 & .359 & .513\\
        \textbf{OrthogonalE}(4$\times$4)   & \underline{.493} & \underline{.446} & \underline{.506} & .578 & .332 & .240 & .363 & .517\\
        \textbf{OrthogonalE}(10$\times$10) & \textbf{.494} & \underline{.446} & \textbf{.508} & .573 & \underline{.334} & \underline{.242} & .367 & \underline{.518}\\
    \clineB{1-9}{2}
    \end{tabular}
    }
\caption{Link prediction accuracy results of two datasets, \textbf{Bold} indicates the best score, and \underline{underline} represents the second-best score. For a fair comparison, we standardized $m$ at 1 for Gram-Schmidt and all OrthogonalE sizes. The entity dimension for WN18RR was set at approximately 500 (for example, 501 for 3$\times$3 blocks to ensure experimental feasibility) and around 1000 for FB15K-237. [$\diamondsuit$]: The results are sourced from \cite{zhang2019quaternion}. For a fair comparison, the results of RotatE, QuatE, HopfE, and DensE are reported without self-adversarial negative sampling, type constraints, semantics, or reciprocals. More baseline results are shown in Appendix~\ref{subsec:baseline_KGE}.}
    \label{tab:main_accuracy_results}
\end{table*}

\subsection{Optimization}
Traditional KGE models train and optimize relations and entities jointly. In contrast, our study aims to achieve more effective optimization of the block-diagonal orthogonal matrices of relation embeddings $ \mathbf{X}_{i} \in \mathbb{R}^{d \times d} $ by separately optimizing relations and entities, utilizing Riemannian optimization for the relation matrix $ \mathbf{e}_{R} \in \mathbb{R}^{n \times n} $ and SGD for the entity matrix $ \mathbf{e}_{V} \in \mathbb{R}^{n \times m} $.

Initially, when optimizing relations, all entity parameters are fixed, rendering the entity embeddings analogous to the function $ f(\cdot) $ in the problem formulated in (\ref{eq:optimization_problem}). Notably, each block-diagonal orthogonal matrix $ \mathbf{X}_{i} $ within the relation embedding $\mathbf{e}_{R}$ optimized by individual Riemannian optimization of eq.~(\ref{eq:exponential_iteration}) using RiemannianAdam \cite{kochurov2020geoopt}, which is a Riemannian version of the popular Adam optimizer \cite{kingma2014adam}. These are then concatenated in a block-diagonal way according to eq.~(\ref{eq:relation_embedding}) to complete the process. After optimizing the relation parameters $ \mathbf{e}_{R} \in \mathbb{R}^{n \times n} $, they are held constant while the entity parameters $ \mathbf{e}_{V} \in \mathbb{R}^{n \times m} $ are optimized using Stochastic Gradient Descent (SGD), specifically employing the Adagrad optimizer \cite{duchi2011adaptive}.

\section{Experiment}

We expect that our proposed OrthogonalE model, employing matrices for entities and block-diagonal orthogonal matrices with Riemannian optimization for relations, will outperform baseline models. Also, we anticipate that OrthogonalE is a general and flexible KGE model and can represent several relation patterns simultaneously. Our goal is to validate these through empirical testing.

\subsection{Experiment Setup}

%.....................................table dataset
\begin{table}[t]
\resizebox{\textwidth}{!}{\renewcommand{\arraystretch}{1.1}
   \centering
   \begin{tabular}{lccccc}
   \clineB{1-6}{2}
   {Dataset} & {Entities} & {Relations} & {Train} & {Validation} & {Test} \\
   \clineB{1-6}{2}
   WN18RR & 40,943 & 11 & 86,835 & 3,034 & 3,134\\
   FB15K-237 & 14,541 & 237 & 272,115 & 17,535 & 20,466 \\
   \clineB{1-6}{2}
   \end{tabular}
   }
\caption{Details of the two datasets.}
\label{tab:datasets}
\end{table}
%....................................table dataset

\paragraph{Dataset.} 
We evaluate our proposed method on two KG datasets, including WN18RR \cite{dettmers2018convolutional} (license: \href{https://www.apache.org/licenses/LICENSE-2.0}{Apache 2.0}), FB15K-237 \cite{toutanova2015observed} (license: CC-BY-4.0). The details of these datasets are shown in Table \ref{tab:datasets}. More detail is given in Appendix~\ref{subsec:experiment_setup}.

\paragraph{Evaluation metrics.} 
To predict the tail entity from a given head entity and relation, we rank the correct tail entity among all possible entities using two established ranking metrics. The first is the mean reciprocal rank (MRR), the average inverse ranking of the correct entities, calculated as $\frac{1}{n} \sum_{i=1}^{n} \frac{1}{\text { Rank }_{i}}$. The second is Hits@$K$ for $K \in \{1,3,10\}$, the frequency of correct entities ranking within the top $K$ positions.

\paragraph{Baselines.} We compare our new model with several classic methods, including TransE \cite{bordes2013translating}, DistMult \cite{yang2014embedding}, ComplEx \cite{trouillon2016complex}, and ConvE \cite{dettmers2018convolutional}. Additionally, we include rotation-based KGE methods such as RotatE \cite{sun2019rotate}, QuatE \cite{zhang2019quaternion}, HopfE \cite{bastos2021hopfe}, and DensE \cite{lu2022dense} as baselines. In addition to these methods and our OrthogonalE$(d \times d)$, we introduce comparative models such as Gram-Schmidt$(d \times d)$ utilizing the Gram-Schmidt process for generating orthogonal matrices and SGD for joint relation-entity training. OrthogonalE further differentiates by employing orthogonal matrices of varying sizes to discuss performance nuances.

\paragraph{Implementation}
The key hyperparameters of our implementation include the learning rate for RiemannianAdam \cite{kochurov2020geoopt} and Adagrad \cite{duchi2011adaptive}, negative sample size, and batch size. To determine the optimal hyperparameters, we performed a grid search using the validation data. More detail refers to Appendix~\ref{subsec:experiment_setup}.

%.....figure in the result
\begin{figure*}[t]
\centering
\begin{subfigure}[b]{1\textwidth}
        \includegraphics[width=\textwidth]{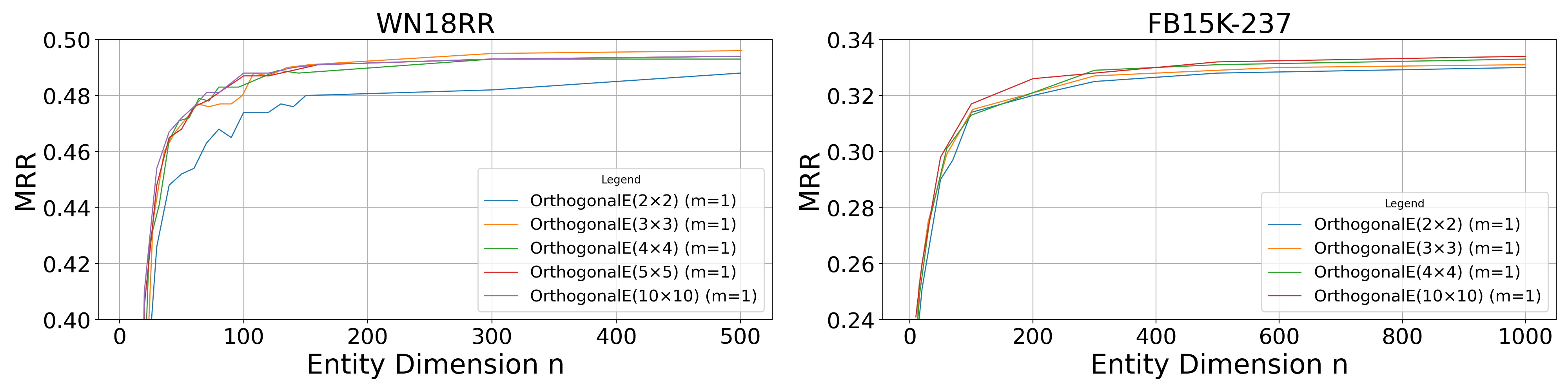}
    \end{subfigure}
\caption{MRR accuracy comparison of OrthogonalE models with different block-diagonal orthogonal matrices across varying entity dimensions ($n\times 1$, where we set $m=1$ to control the entity shape as a single entity vector) on WN18RR and FB15K-237.}
\label{fig:WN18RR_FB15K-237_small_block}
\end{figure*}
%......figure in the result

\subsection{Results}
We first analyzed the overall accuracy for all baseline models and OrthogonalE, then separately examined the impacts of block-diagonal Orthogonal matrices, Riemannian Optimization for relations, and entity matrices on the model from various experimental results. Finally, we utilize several relation histograms to verify our model can capture these relation patterns.

\subsubsection{Overall Accuracy}
Table \ref{tab:main_accuracy_results} presents link prediction accuracies for the WN18RR and FB15K-237 datasets. The OrthogonalE model demonstrates superior performance in the WN18RR dataset and achieves results on the FB15K-237 dataset that are only marginally lower than those of HopfE \cite{bastos2021hopfe}, outperforming all other compared models, highlighting its superior representational ability by employing matrices for entities and block-diagonal orthogonal matrices with Riemannian optimization for relations. Moreover, the OrthogonalE model with 2$\times$2 and 3$\times$3 configurations yields significantly better performance than the corresponding sizes of the Gram-Schmidt method, and notably exceeds RotatE and QuatE, respectively, showcasing the enhanced efficacy of the KGE model. Finally, since the WN18RR and FB15K-237 datasets are relatively small, the performance differences among OrthogonalE models with ($2 \times 2$), ($3 \times 3$), ($4 \times 4$), and ($10 \times 10$) are not significant when using sufficient dimensions (WN18RR: 500, FB15K-237: 1000). We will discuss the performance at different dimensions in detail in Section~\ref{subsec:Orthogonal_Relation}.

\subsubsection{Block-diagonal Orthogonal matrices for generality} \label{subsec:Orthogonal_Relation}

To prove the generality of OrthogonalE, we design experiments to compare the MRR accuracy of OrthogonalE models with different block-diagonal orthogonal matrices across varying entity dimensions ($n\times 1$, where we set $m=1$ to control the entity shape as a single entity vector) on WN18RR and FB15K-237. The results are shown in Fig.~\ref{fig:WN18RR_FB15K-237_small_block} and the explanation is provided as follows.

An initial dataset analysis reveals WN18RR has 40,943 entities with just 11 relations (about 3,722 entities per relation), while FB15K-237 includes 14,541 entities and 237 relations (around 61 entities per relation). This implies that WN18RR requires a more sophisticated representation capability compared to FB15K-237.

Our results (Fig.~\ref{fig:WN18RR_FB15K-237_small_block}) confirm our dataset analysis. For WN18RR, the performance is similar for block sizes from 3$\times$3 to 10$\times$10, all outperforming 2$\times$2 blocks, showcasing 2$\times$2 blocks are not enough for its relation representation. However, for FB15K-237, performance is stable across all block sizes, indicating 2$\times$2 blocks are enough for its relations representation. These results show WN18RR requires more complex blocks for adequate representation, and illustrate that the OrthogonalE model is general, which can adapt to datasets of various complexities by adjusting the dimension $d$ of the block-diagonal matrices.

%.....figure in the result
\begin{figure}[t] 
\centering
\begin{subfigure}[b]{1\textwidth}
        \includegraphics[width=\textwidth]{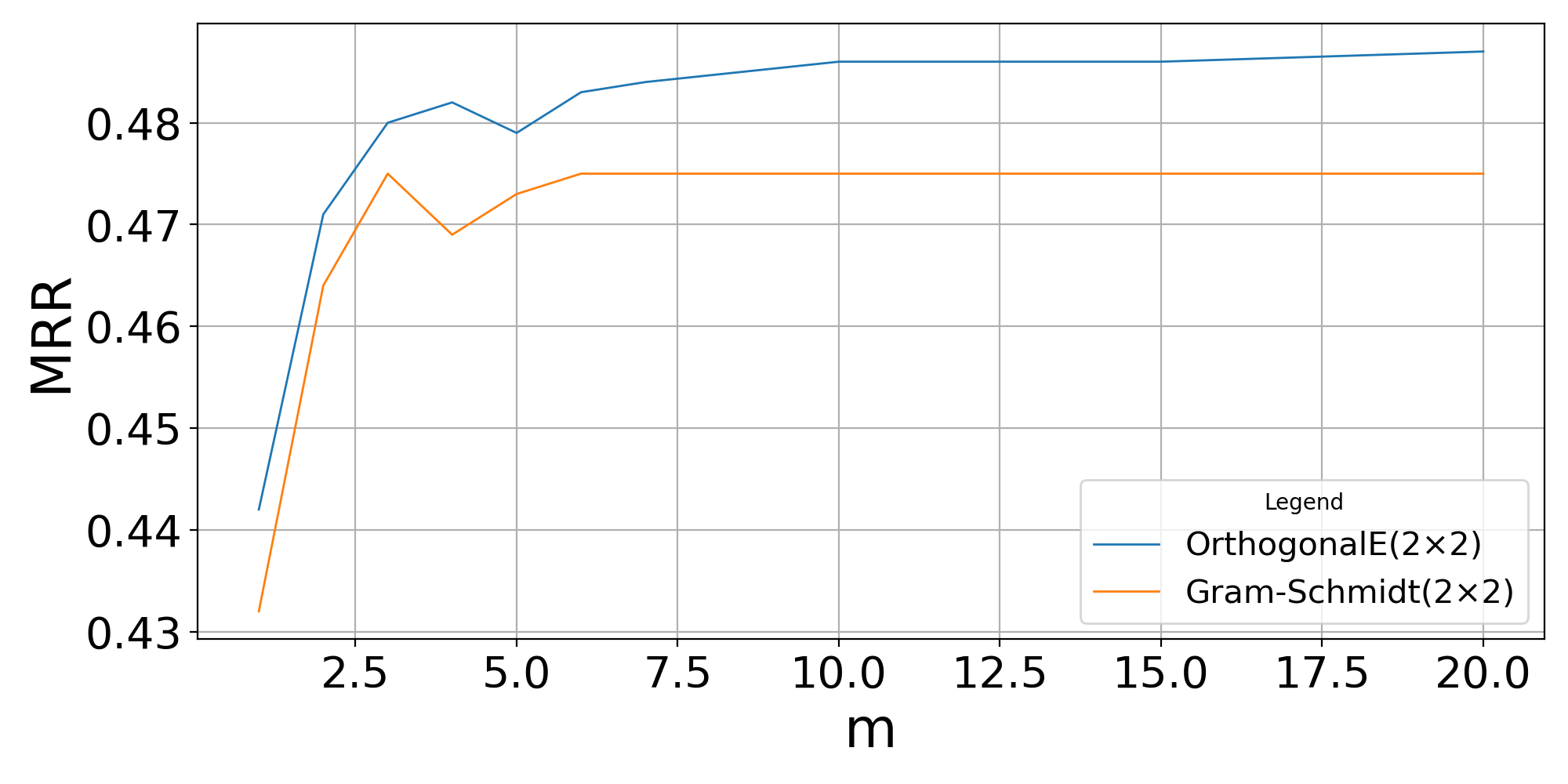}
    \end{subfigure}
\caption{MRR accuracy comparison of OrthogonalE(2$\times$2) and Gram-Schmidt(2$\times$2) models across varying entity dimensions ($m$) with fixed relation matrix (40$\times$40) on WN18RR.}
\label{fig:orthogonal_schmidt}
\end{figure}

%......figure in the result

%......figure in the result
\begin{figure}[t]
\centering
\begin{subfigure}[b]{1\textwidth}
        \includegraphics[width=\textwidth]{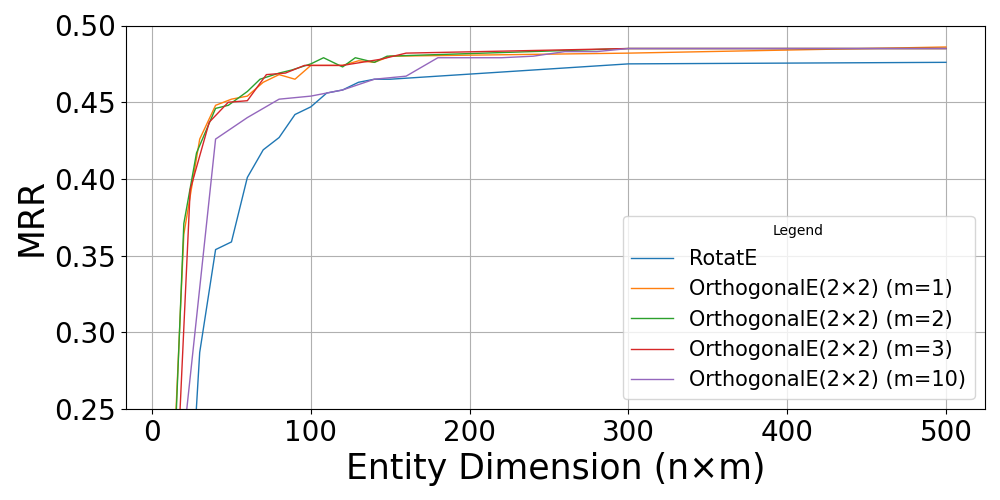}
    \end{subfigure}
\caption{MRR accuracy comparison of RotatE and OrthogonalE(2$\times$2) models across varying entity dimensions ($n\times m$) on WN18RR.}
\label{fig:RotatE_Orthogonal}
\end{figure}

%......figure in the result

%.....figure in the result
\begin{figure}[t]
\centering
\begin{subfigure}[b]{1\textwidth}
        \includegraphics[width=\textwidth]{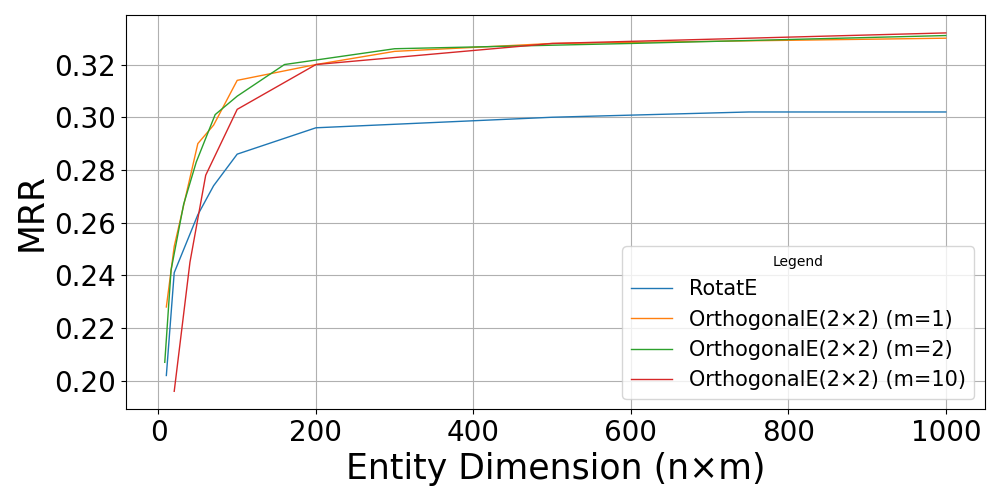}
    \end{subfigure}
\caption{MRR accuracy comparison of RotatE and OrthogonalE(2$\times$2) models across varying entity dimensions ($n\times m$) on FB15K-237.}
\label{fig:RotatE_OrthogonalE_FB237}
\end{figure}

%......figure in the result

%.....figure in the result
\begin{figure}[t]
\centering
\begin{subfigure}[b]{1\textwidth}
        \includegraphics[width=\textwidth]{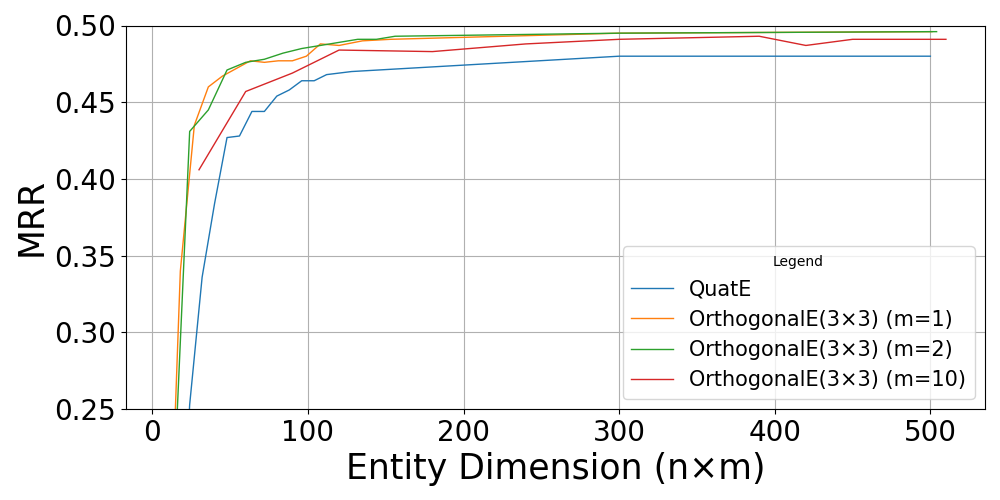}
    \end{subfigure}
\caption{MRR accuracy comparison of QuatE and OrthogonalE(3$\times$3) models across varying entity dimensions ($n\times m$) on WN18RR.}
\label{fig:QuatE_Orthogonal}
\end{figure}

%......figure in the result

%.....figure in the result
\begin{figure*}[t]
\centering
\begin{subfigure}[b]{1\textwidth}
        \includegraphics[width=\textwidth]{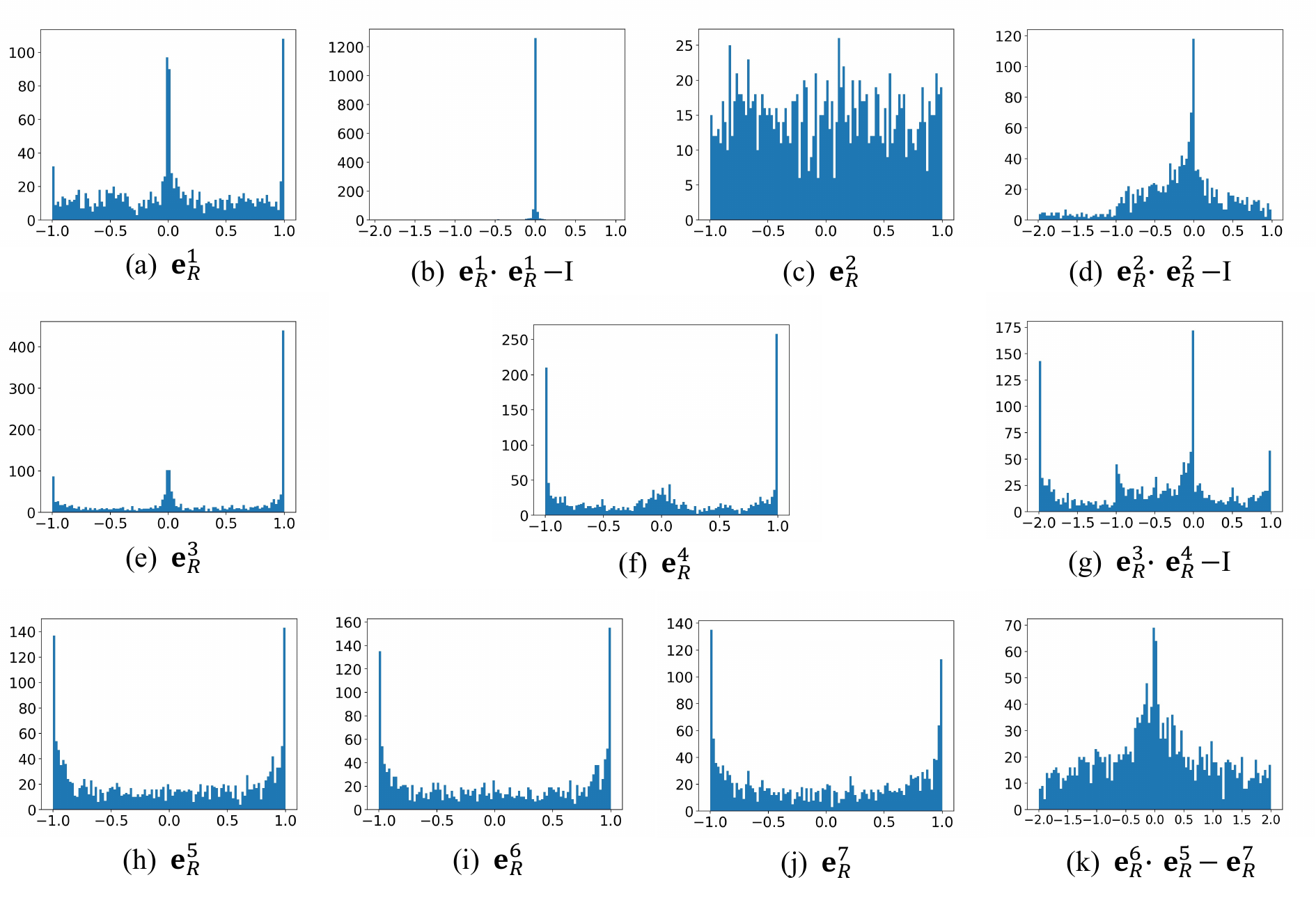}
    \end{subfigure}
\caption{Histograms of relation embeddings for different relation patterns, where $ \mathbf{e}_{R}^{1} $ represents \texttt{\_similar\_to}, $\mathbf{e}_{R}^{2}$ represents \texttt{\_member\_of\_domain\_region}, $ \mathbf{e}_{R}^{3} $ represents \texttt{/film/film/genre}, $ \mathbf{e}_{R}^{4} $ represents \texttt{/media\_common/netflix\_genre/titles}, $ \mathbf{e}_{R}^{5} $ represents \texttt{/location/administrative\_division/country}, $ \mathbf{e}_{R}^{6} $ represents \texttt{/location/hud\_county\_place/place}, and $ \mathbf{e}_{R}^{7} $ represents \texttt{/base/aareas/schema/administrative\_area/capital}. From the WN18RR dataset, we select $ \mathbf{e}_{R}^{1} $ and $ \mathbf{e}_{R}^{2} $ to represent \textit{Symmetry} and \textit{Antisymmetry}, respectively, and obtain their relation embeddings using the OrthogonalE(3$\times$3) model with $n$=500 and $m$=1. Similarly, from the FB15K-237 dataset, we select $ \mathbf{e}_{R}^{3} $, $ \mathbf{e}_{R}^{4} $, and $ \mathbf{e}_{R}^{5} $, $ \mathbf{e}_{R}^{6} $, $ \mathbf{e}_{R}^{7} $ as representations for \textit{Inversion} and \textit{Composition}, respectively, and acquire their relation embeddings under the OrthogonalE(2$\times$2) model with $n$=1000 and $m$=1.}

\label{fig:relation_pattern_histgram}
\end{figure*}

%......figure in the result

\subsubsection{Riemannian Optimization for relations}
Fig.~\ref{fig:orthogonal_schmidt} compares MRR accuracies of OrthogonalE (2$\times$2) and Gram-Schmidt (2$\times$2) across entity dimensions ($m$) with a constant relation matrix (40$\times$40) on WN18RR, assessing the efficacy of orthogonal optimization beyond the Gram-Schmidt method for block-diagonal orthogonal matrices. The result demonstrates that OrthogonalE's Riemannian optimization significantly exceeds Gram-Schmidt, underscoring its necessity. 
In addition, OrthogonalE can solve the problem that Gram-Schmidt model will face the gradient vanishing during the training process.

\subsubsection{Entity matrix for flexibility} \label{subsec:Entity}
In OrthogonalE, we maintained a constant entity dimension ($n\times m$) while varying $m$ to assess the impact of entity shape. To prove the flexibility of OrthogonalE, we design experiments to compare the MRR accuracies of RotatE with OrthogonalE (2$\times$2) and QuatE and OrthogonalE (3$\times$3) over different fixed entity dimensions $n\times m$ in WN18RR and FB15K-237 dataset.

From the results on WN18RR in Fig.~\ref{fig:RotatE_Orthogonal}, OrthogonalE models with $m=1$, $2$, or $3$ perform similarly and better than $m=10$, and all significantly outperform RotatE across dimensions. Notably, their relation parameter count is $1/m$ of RotatE's, which is shown in Table.~\ref{tab:parameter_count}. These results demonstrate OrthogonalE's efficacy in saving relation parameters while outperforming RotatE, highlighting our model's flexibility in controlling entity dimension through variable $m$ without unnecessarily increasing relation size. 

Furthermore, Fig.~\ref{fig:RotatE_OrthogonalE_FB237} shows comparison of RotatE and OrthogonalE (2$\times$2) in FB15K-237. Fig.~\ref{fig:QuatE_Orthogonal} compare MRR accuracies of QuatE and OrthogonalE (3$\times$3) on WN18RR. The experimental results, consistent with those discussed in the previous paragraph.

Furthermore, for OrthogonalE(2$\times$2) on WN18RR dataset, Fig.~\ref{fig:orthogonal_schmidt} result (with $m=7$ yielding MRR=0.483) suggests that a relation matrix of 40$\times$40 (20 parameters), compared to a dimension of 500$\times$500 (250 parameters) in Table~\ref{tab:main_accuracy_results}, can achieve comparably high performance, thus demonstrating that entity matrix method significantly reduces the need for excessive relation parameters.

\subsubsection{Relation Pattern}
Following the proof of relation patterns in Appendix~\ref{sec:proof_relation_pattern}, Fig.~\ref{fig:relation_pattern_histgram} shows histograms of relation embeddings for different relation patterns. We provide several examples of relation patterns and a discussion of \textit{non-commutative composition} in Appendix~\ref{appendix:relation_pattern}.

\paragraph{Symmetry and Antisymmetry}
In OrthogonalE, the \textit{symmetry} relation pattern is encoded when the $ \mathbf{e}_{R} $ embedding satisfies $\mathbf{e}_{R} \cdot \mathbf{e}_{R} = \mathrm{I}$, in accordance with eq.~(\ref{equ:symmetery_proof}). Figs.~\ref{fig:relation_pattern_histgram}(a) and (b) illustrate the embeddings of $ \mathbf{e}_{R}^{1} $ and $\mathbf{e}_{R}^{1} \cdot \mathbf{e}_{R}^{1} - \mathrm{I}$, respectively. From Fig.~\ref{fig:relation_pattern_histgram}(b), we observe that nearly all values are concentrated around 0, thereby indicating that OrthogonalE's relations exhibit \textit{symmetry} properties. Correspondingly, the multitude of nonzero values in Fig.~\ref{fig:relation_pattern_histgram}(d) indicates that OrthogonalE's relations also can represent \textit{antisymmetry} properties.

\paragraph{Inversion}
The \textit{inversion} relation pattern is encoded when the $ \mathbf{e}_{R}^{3} $ and $ \mathbf{e}_{R}^{4} $ satisfies $\mathbf{e}_{R}^{3} \cdot \mathbf{e}_{R}^{4} = \mathrm{I}$, according to eq.~(\ref{equ:inversion_proof}). Even though $ \mathbf{e}_{R}^{3} $ and $ \mathbf{e}_{R}^{4} $ are responsible for additional relation patterns, which results in a cluster of values around $-2$ in Fig~\ref{fig:relation_pattern_histgram} (g), the majority of values still converge towards or equal 0. This suggests that OrthogonalE's relations have the \textit{inversion} property.

\paragraph{Composition}
The \textit{composition} relation pattern is encoded when the $ \mathbf{e}_{R}^{5} $, $ \mathbf{e}_{R}^{6} $, and $ \mathbf{e}_{R}^{7} $ embedding satisfy $\mathbf{e}_{R}^{6} \cdot \mathbf{e}_{R}^{5} = \mathbf{e}_{R}^{7}$, in accordance with eq.~(\ref{equ:composition_proof}).
The majority of data in Fig.~\ref{fig:relation_pattern_histgram} (k) converge towards or are equal to 0, indicating that OrthogonalE's relations can represent the \textit{composition} relation pattern.

\section{Conclusion}
In this study, we propose the OrthogonalE model to acquire a flexible and general KGE model with employing matrices for entities and block-diagonal orthogonal matrices with Riemannian optimization for relations. Experimental results indicate that our new KGE model offers generality and flexibility, captures several relation patterns, and outperforms SoTA rotation-based KGE models while substantially reducing the number of relation parameters.

\clearpage

%.............................Limitations

\section*{Limitations}
Even though the block-diagonal orthogonal relation with Riemannian optimization makes KGE models more general and improves their performance, the computation of exponential retraction in the orthogonal manifold for Riemannian optimization is costly. In practical model training, with the same entity dimension, our OrthogonalE (2$\times$2) training time is 4 times longer than that of RotatE. In future research directions, we will continue to explore this limitation, such as by employing the landing algorithm \cite{ablin2022fast} for retraction on orthogonal manifolds to reduce computational complexity.

\section*{Ethics Statement}
This study complies with the \href{https://www.aclweb.org/portal/content/acl-code-ethics}{ACL Ethics Policy}.

\section*{Acknowledgements}

This work was supported by JST SPRING, Grant Number JPMJSP2110 (YZ).
This study was partially supported by JSPS KAKENHI 22H05106, 23H03355, and JST CREST JPMJCR21N3 (HS).
Additionally, we extend our gratitude to Kenji Fukumizu and Kei Hirose for engaging in insightful discussions and to the anonymous reviewers for their constructive feedback.

%.............................References
\bibliography{references}

%.............................Appendix
\appendix
\section{Appendix}

\begin{figure*}[t]
\centering
\begin{subfigure}[b]{1\textwidth}
        \includegraphics[width=\textwidth]{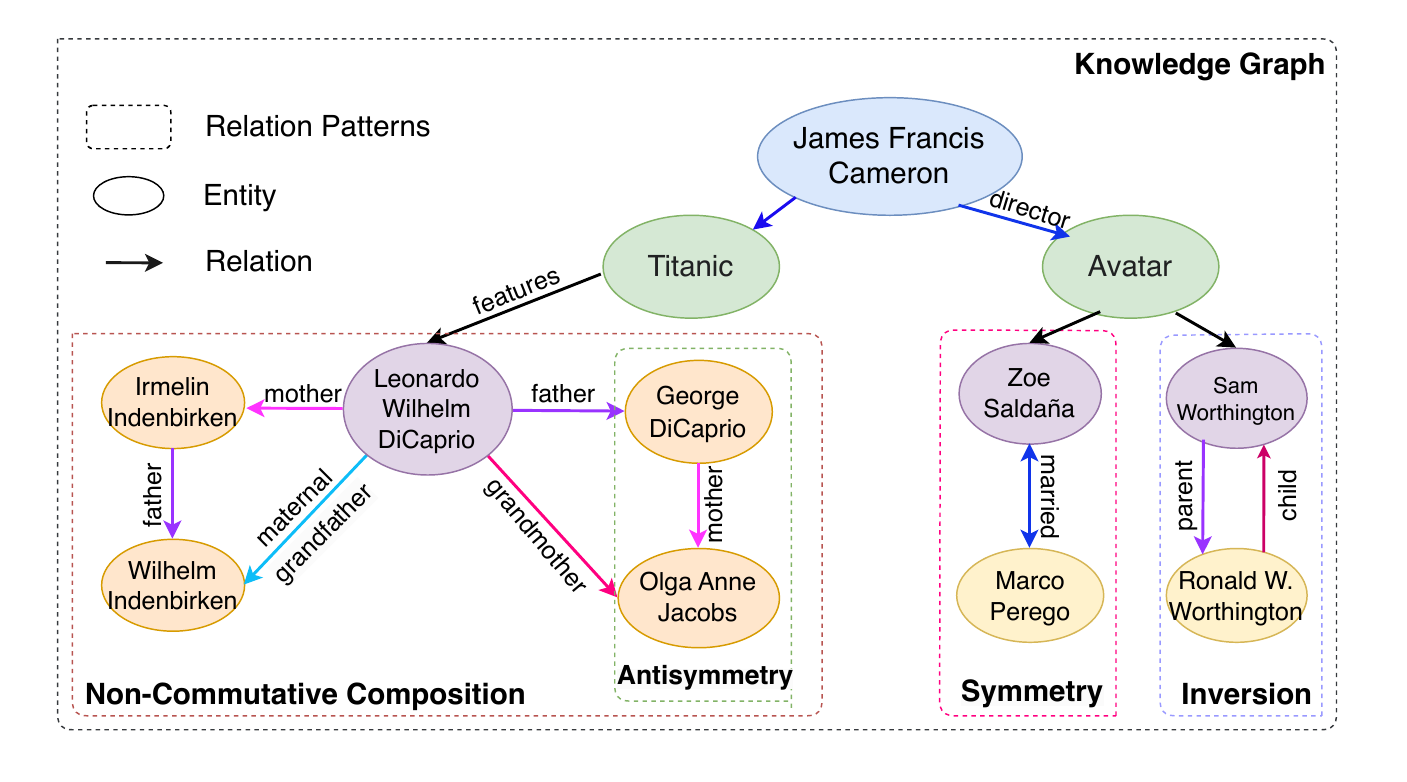}
    \end{subfigure}
\caption{Toy examples for \textit{symmetry}, \textit{antisymmetry}, \textit{inversion}, and \textit{non-commutative composition} relation patterns.}
\label{fig:relation_pattern_examples}
\end{figure*}

%.....figure in the result
\begin{figure*}[t]
\centering
\begin{subfigure}[b]{1\textwidth}
        \includegraphics[width=\textwidth]{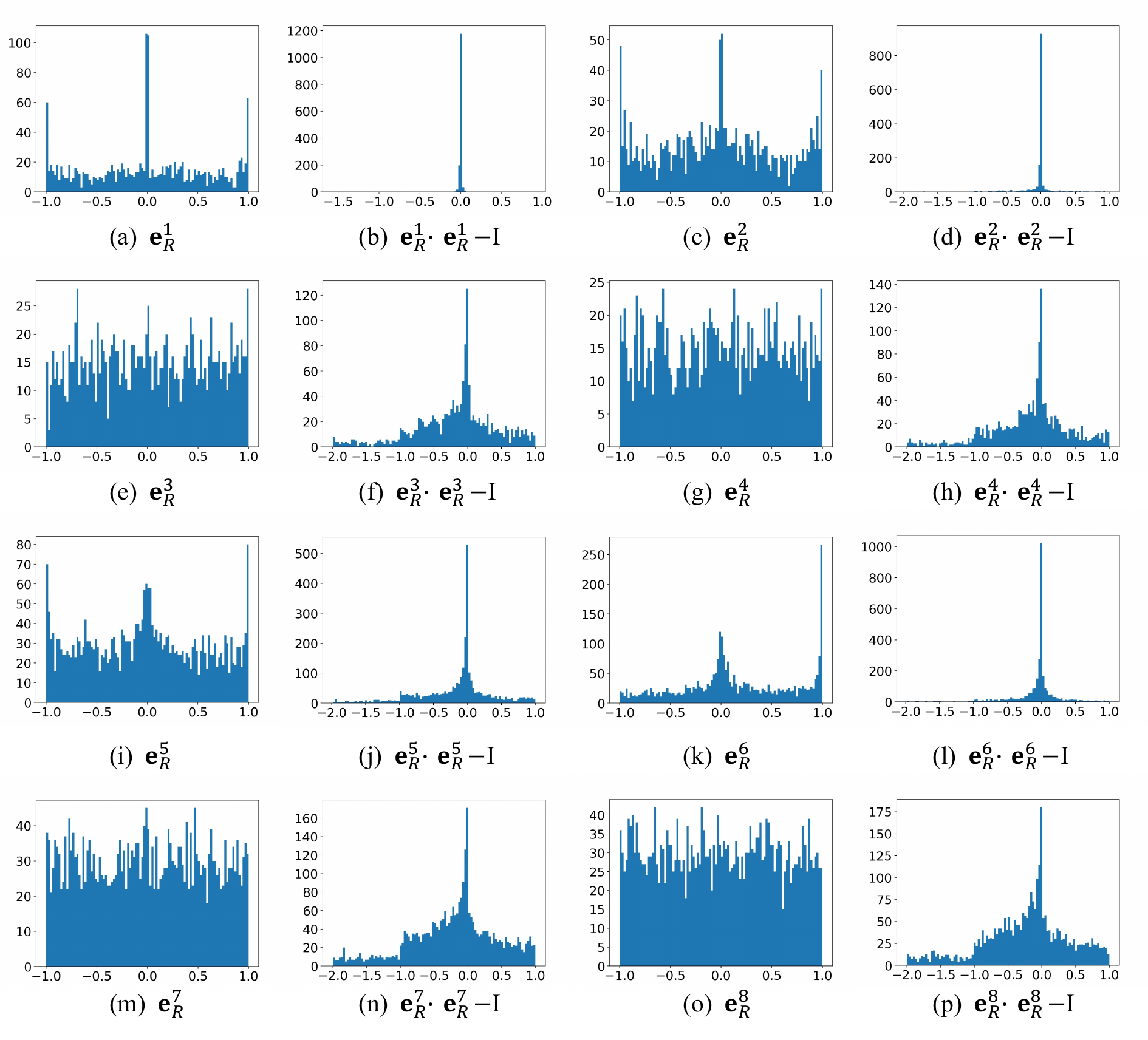}
    \end{subfigure}
\caption{Histograms of relation embeddings for \textit{symmetry} and \textit{antisymmetry} relation patterns, where $ \mathbf{e}_{R}^{1} $ represents \texttt{\_derivationally\_related\_form}, $\mathbf{e}_{R}^{2}$ represents \texttt{\_instance\_hypernym}, $ \mathbf{e}_{R}^{3} $ represents \texttt{\_also\_see}, $ \mathbf{e}_{R}^{4} $ represents \texttt{\_verb\_group}, $ \mathbf{e}_{R}^{5} $ represents \texttt{/media\_common/netflix\_genre/titles}, $ \mathbf{e}_{R}^{6} $ represents \texttt{/film/film/genre}, $ \mathbf{e}_{R}^{7} $ represents \texttt{/award/award\_category/category\_of	}, and $ \mathbf{e}_{R}^{8} $ represents \texttt{/people/person/gender}. From the WN18RR dataset, we select $ \mathbf{e}_{R}^{1} $, $ \mathbf{e}_{R}^{2} $ $ \mathbf{e}_{R}^{3} $, $ \mathbf{e}_{R}^{4} $and to represent \textit{Symmetry} and \textit{Antisymmetry}, respectively, and obtain their relation embeddings using the OrthogonalE(3$\times$3) model with $n$=501 and $m$=1. Similarly, from the FB15K-237 dataset, we select $ \mathbf{e}_{R}^{5} $, $ \mathbf{e}_{R}^{6} $, and $ \mathbf{e}_{R}^{7} $, $ \mathbf{e}_{R}^{8} $ as representations for \textit{symmetry} and \textit{antisymmetr}, respectively, and acquire their relation embeddings under the OrthogonalE(3$\times$3) model with $n$=999 and $m$=1.}

\label{fig:more_symmetry}
\end{figure*}

%.....figure in the result
\begin{figure*}[t]
\centering
\begin{subfigure}[b]{1\textwidth}
        \includegraphics[width=\textwidth]{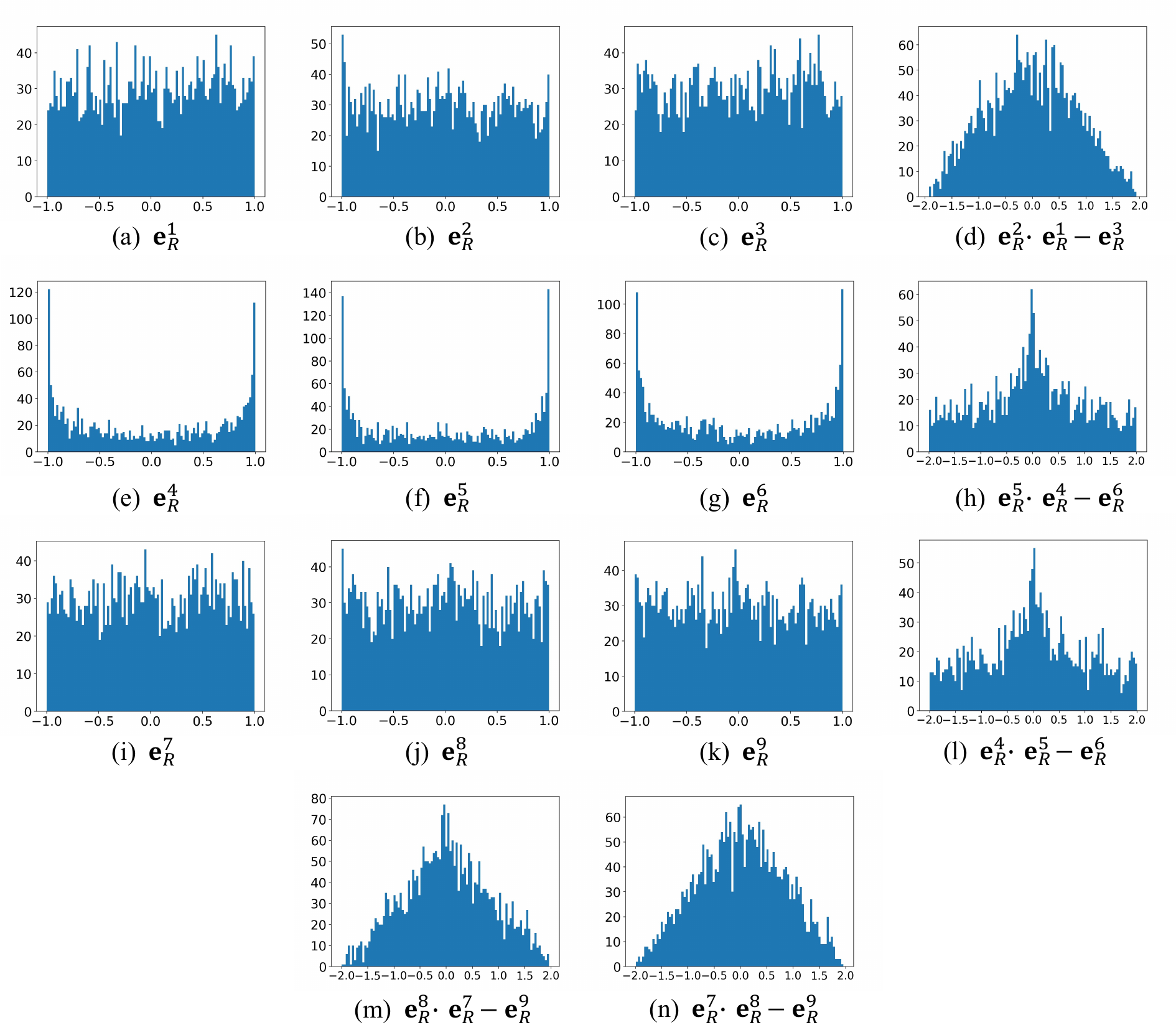}
    \end{subfigure}
\caption{Histograms of relation embeddings for \textit{composition} relation patterns, where $ \mathbf{e}_{R}^{1} $ represents \texttt{/location/administrative\_division/country}, $\mathbf{e}_{R}^{2}$ represents \texttt{/location/hud\_county\_place/place}, $ \mathbf{e}_{R}^{3} $ represents \texttt{/base/aareas/schema/administrative\_area/capital}, $ \mathbf{e}_{R}^{4} $ represents \texttt{/award/award\_nominee/award\_nominations./award/award\_nomination/nominated\_for}, $ \mathbf{e}_{R}^{5} $ represents \texttt{/award/award\_category/winners./award/award\_honor/award\_winner}, and $ \mathbf{e}_{R}^{6} $ represents \texttt{/award/award\_category/nominees./award/award\_nomination/nominated\_for}. $ \mathbf{e}_{R}^{7} $, $ \mathbf{e}_{R}^{8} $, and $ \mathbf{e}_{R}^{9} $ have the same relational meanings as $ \mathbf{e}_{R}^{4} $, $ \mathbf{e}_{R}^{5} $, and $ \mathbf{e}_{R}^{6} $, respectively, the difference lies in that the former are relations within the OrthogonalE(3$\times$3) model, while the latter are from the OrthogonalE(2$\times$2) model. All these relations are selected from the FB15K-237 dataset. $ \mathbf{e}_{R}^{1} $, $ \mathbf{e}_{R}^{2} $, $ \mathbf{e}_{R}^{3} $, $ \mathbf{e}_{R}^{7} $, $ \mathbf{e}_{R}^{8} $, and $ \mathbf{e}_{R}^{9} $ are relation embeddings under the OrthogonalE(3$\times$3) model with $n$=999 and $m$=1, while $ \mathbf{e}_{R}^{4} $, $ \mathbf{e}_{R}^{5} $, and $ \mathbf{e}_{R}^{6} $ are relation embeddings under the OrthogonalE(2$\times$2) model with $n$=1000 and $m$=1}

\label{fig:more_composition}
\end{figure*}

%..............hyperparameter

\begin{table*}[t]
\centering

\begin{tabular}{l|clcccc}
\clineB{1-6}{2}
{Dataset} & model & {lr-entity} & {lr-relation} & {optimizer} & {negative samples} \\
\clineB{1-6}{2}
\multirow{4}{*}{{WN18RR(dim=500)}} 
& TransE & 0.001 & - & Adam & 300 \\
& RotatE & 0.1 & - & Adagrad & 300 \\
& QuatE & 0.2 & - & Adagrad & 300\\
& OrthogonalE(2$\times$2) & 0.2 & 0.02 & - & 300\\ 
& OrthogonalE(3$\times$3) & 0.2 & 0.02 & - & 300\\ 

\clineB{1-6}{2}
\multirow{4}{*}{{FB15k-237(dim=1000)}} 
& TransE & 0.05 & - & Adam & 300 \\
& RotatE & 0.1 & - & Adagrad & 300 \\
& QuatE & 0.2 & - & Adagrad & 300\\
& OrthogonalE(2$\times$2) & 0.5 & 0.06 & - & 300\\ 
& OrthogonalE(3$\times$3) & 0.5 & 0.06 & - & 300\\ 

\clineB{1-6}{2}
\end{tabular}
\caption{Best hyperparameters of our approach and several composite models. In the table, the lr-entity values corresponding to TransE, RotatE, and QuatE refer to the learning rate for the entire model. For the OrthogonalE model, we employ RiemannianAdam for relation optimization and Adagrad for entity optimization, as detailed in the Approach section.} 
\label{tab:hyperparameter}
\end{table*}
%..............hyperparameter

\subsection{More information about Experiment setup}  \label{subsec:experiment_setup}
\paragraph{Dataset}
WN18RR is a subset of WN18 \cite{dettmers2018convolutional} which is contained in WordNet \cite{miller1995wordnet}. FB15K-237 is a subset of FB15K, which is a subset of Freebase \cite{bollacker2008freebase}, a comprehensive KG including data about common knowledge.
All three datasets were designed for KGE, and we employ them for KGE tasks, and all three datasets have no individual people or offensive content.

\paragraph{Implementation}
The training of models was carried out on two A6000 GPUs, which boasts 48GB of memory. Specifically, for the OrthogonalE model and its related flexible versions, the training durations were roughly 5 hour for the WN18RR dataset, 30 hours for FB15K-237. Our experiments were facilitated by leveraging \href{https://pytorch.org}{PyTorch} and \href{https://numpy.org}{Numpy} as essential tools. Furthermore, We use \href{https://chat.openai.com/#}{ChatGPT} in our paper writing. Finally, we obtain results by selecting the maximum values from three random seeds for Table~\ref{tab:main_accuracy_results} and using a single run for other results.

\subsection{Proof of Relation Patterns}  \label{sec:proof_relation_pattern}

OrthogonalE is capable of representing the four kinds of relational patterns: \textit{Symmetry}, \textit{Antisymmetry},  \textit{Inversion}, and \textit{Non-commutative Composition}. We present the following propositions to formalize this capability:

\paragraph{Proposition 1.} \textit{OrthogonalE can represent Symmetry relation pattern.}

\textit{Proof.} If $ (\mathbf{e}_{H}, \mathbf{e}_{R}, \mathbf{e}_{T}) \in \mathcal{E} $ and $ (\mathbf{e}_{T}, \mathbf{e}_{R}, \mathbf{e}_{H}) \in \mathcal{E} $, we have
\begin{align}
\begin{split}
    &\mathbf{e}_{R} \cdot \mathbf{e}_{H} = \mathbf{e}_{T}  \land \mathbf{e}_{R} \cdot \mathbf{e}_{T} = \mathbf{e}_{H} \\
                                                       & \Rightarrow \mathbf{e}_{R} \cdot \mathbf{e}_{R} = \mathrm{I} \\
                                                       & \Rightarrow \mathbf{X}_{i} \cdot \mathbf{X}_{i} = \mathrm{I} \\ \label{equ:symmetery_proof}
\end{split}
\end{align}

\paragraph{Proposition 2.} \textit{OrthogonalE can represent Antisymmetry relation pattern.}

\textit{Proof.} If $ (\mathbf{e}_{H}, \mathbf{e}_{R}, \mathbf{e}_{T}) \in \mathcal{E} $ and $ (\mathbf{e}_{T}, \mathbf{e}_{R}, \mathbf{e}_{H}) \notin \mathcal{E} $, we have
\begin{align}
\begin{split}
    &\mathbf{e}_{R} \cdot \mathbf{e}_{H} = \mathbf{e}_{T}  \land \mathbf{e}_{R} \cdot \mathbf{e}_{T} \neq \mathbf{e}_{H} \\
                                                       & \Rightarrow \mathbf{e}_{R} \cdot \mathbf{e}_{R} \neq \mathrm{I} \\
                                                       & \Rightarrow \mathbf{X}_{i} \cdot \mathbf{X}_{i} \neq \mathrm{I} \\
\end{split}
\end{align}

\paragraph{Proposition 3.} \textit{OrthogonalE can represent Inversion relation pattern.}

\textit{Proof.}  If $ (\mathbf{e}_{H}, \mathbf{e}_{R}^{1}, \mathbf{e}_{T}) \in \mathcal{E} $ and $ (\mathbf{e}_{T}, \mathbf{e}_{R}^{2}, \mathbf{e}_{H}) \in \mathcal{E} $, we have
\begin{align}
\begin{split}
    &\mathbf{e}_{R}^{1} \cdot \mathbf{e}_{H} = \mathbf{e}_{T}  \land \mathbf{e}_{R}^{2} \cdot \mathbf{e}_{T} = \mathbf{e}_{H} \\
                                                       & \Rightarrow \mathbf{e}_{R}^{1} \cdot \mathbf{e}_{R}^{2} = \mathrm{I} \\
                                                       & \Rightarrow \mathbf{X}_{i}^{1} \cdot \mathbf{X}_{i}^{2} = \mathrm{I} \\ \label{equ:inversion_proof}
\end{split}
\end{align}

\paragraph{Proposition 4.} \textit{OrthogonalE can represent Non-commutative Composition relation pattern.}

\textit{Proof.}  If $ (\mathbf{e}_{H}, \mathbf{e}_{R}^{1}, \mathbf{e}_{T}) \in \mathcal{E} $ and $ (\mathbf{e}_{T}, \mathbf{e}_{R}^{2}, \mathbf{e}_{P}) \in \mathcal{E} $ and $ (\mathbf{e}_{H}, \mathbf{e}_{R}^{3}, \mathbf{e}_{P}) \in \mathcal{E} $, we have
\begin{align}
\begin{split}
    &\mathbf{e}_{R}^{1} \! \cdot \!\mathbf{e}_{H} = \mathbf{e}_{T} \land \mathbf{e}_{R}^{2} \! \cdot \! \mathbf{e}_{T} = \mathbf{e}_{P} \land \mathbf{e}_{R}^{3} \! \cdot \! \mathbf{e}_{H} \! = \mathbf{e}_{P} \\
                                              & \Rightarrow \mathbf{e}_{R}^{2} \cdot \mathbf{e}_{R}^{1} = \mathbf{e}_{R}^{3} \\
                                              & \Rightarrow \mathbf{X}_{i}^{2} \cdot \mathbf{X}_{i}^{1} = \mathbf{X}_{i}^{3} \\  \label{equ:composition_proof}
\end{split}
\end{align}
\begin{align}
    \mathbf{X}_{i} \in \mathbb{R}^{d \times d}  \left\{\begin{array}{l}
    \text{is Commutative} , \text { if } d = 2 \\
    \text{is Non-commutative}, \text { if } d > 2
    \end{array}\right.   \label{equ:non_commutative}
\end{align}

The property of non-commutative composition dictates that the sequence of $ \mathbf{X}_{i}^{1} $ and $ \mathbf{X}_{i}^{2} $ cannot be exchanged. Given that $\mathbf{X}_{i} \in \mathbb{R}^{d \times d} $ represents an orthogonal matrix, we consider two situations. In the first scenario, when $d = 2$, $ \mathbf{X}_{i} $ is a special case corresponding to a 2-dimensional rotation matrix, analogous to the RotatE \cite{sun2019rotate}, and is therefore commutative, not exhibiting non-commutative composition. However, for $ d > 2 $, $ \mathbf{X}_{i} $ is non-commutative, thus can represent non-commutative composition relation pattern.

To gain a clearer understanding of the proof process, we use symmetry as an illustrative example to introduce the proof section, specifically referring to equation~\ref{equ:symmetery_proof} in the paper. Initially, we assume that if a relation $\mathbf{e}_{R}$ in the OrthogonalE model exhibits the property of symmetry, then we can identify two related KG triples: $ (\mathbf{e}_{H}, \mathbf{e}_{R}, \mathbf{e}_{T}) \in \mathcal{E} $ and $ (\mathbf{e}_{T}, \mathbf{e}_{R}, \mathbf{e}_{H}) \in \mathcal{E} $. For instance, $\mathbf{e}_{H}$ ($\mathbf{e}_{R}$: is similar to) $\mathbf{e}_{T}$ and $\mathbf{e}_{T}$ ($\mathbf{e}_{R}$: is similar to) $\mathbf{e}_{H}$. Since both triples are trained by the OrthogonalE model, they adhere to the OrthogonalE equation (as depicted in Fig.~\ref{fig:approach_OrthogonalE}). Consequently, we can derive that $\mathbf{e}_{R} \cdot \mathbf{e}_{H} = \mathbf{e}_{T}$ and $\mathbf{e}_{R} \cdot \mathbf{e}_{T} = \mathbf{e}_{H}$. By combining and simplifying these two equations, we can conclude that $\mathbf{e}_{R} \cdot \mathbf{e}_{R} = \mathrm{I}$ (Identity matrix). This means that if we can identify such an $\mathbf{e}_{R}$ that satisfies $\mathbf{e}_{R} \cdot \mathbf{e}_{R} = \mathrm{I}$, it demonstrates that the OrthogonalE model can capture the symmetry relation pattern. For $\mathbf{e}_{R} \cdot \mathbf{e}_{R} = \mathrm{I}$, we understand that $\mathbf{e}_{R}$ is composed of several block-diagonal orthogonal matrices $\mathbf{X}_{i}$, as shown in Fig.~\ref{fig:approach_OrthogonalE}). Ultimately, this reduces to finding $\mathbf{X}_{i} \cdot \mathbf{X}_{i} = \mathrm{I}$ to satisfy $\mathbf{e}_{R} \cdot \mathbf{e}_{R} = \mathrm{I}$. We can identify corresponding orthogonal matrices $\mathbf{X}_{i}$ that satisfy $\mathbf{X}_{i} \cdot \mathbf{X}_{i} = \mathrm{I}$, which demonstrates that OrthogonalE can fulfill the property of symmetry.

\subsection{More experiments on relation pattern} \label{appendix:relation_pattern}
\paragraph{Symmetry and Antisymmetry}
Fig.~\ref{fig:more_symmetry} shows histograms of additional examples of relation embeddings for \textit{symmetry} and \textit{antisymmetry} relation patterns. Furthermore, it displays examples of two \textit{symmetry} and two \textit{antisymmetry} relations from both the WN18RR and FB15K-237 datasets.

\paragraph{Composition}
Firstly, we add $ \mathbf{e}_{R}^{1} $, $ \mathbf{e}_{R}^{2} $, and $ \mathbf{e}_{R}^{3} $ from OrthogonalE(3$\times$3) for comparison with the three composition relations in Fig.~\ref{fig:relation_pattern_histgram} from OrthogonalE(2$\times$2). From Fig.~\ref{fig:more_composition}(a, b, c, d), we observe that OrthogonalE(2$\times$2) performs better in composition than OrthogonalE(3$\times$3).

Secondly, we aim to explore more about the \textit{non-commutative composition} relation pattern, so we select $ \mathbf{e}_{R}^{4} $, $ \mathbf{e}_{R}^{5} $, and $ \mathbf{e}_{R}^{6} $, three \textit{non-commutative composition} relations, for our study. Notably, $ \mathbf{e}_{R}^{7} $, $ \mathbf{e}_{R}^{8} $, and $ \mathbf{e}_{R}^{9} $ share the same relational meanings as $ \mathbf{e}_{R}^{4} $, $ \mathbf{e}_{R}^{5} $, and $ \mathbf{e}_{R}^{6} $, respectively, with the distinction that the former are relations within OrthogonalE(3$\times$3), while the latter are within OrthogonalE(2$\times$2). Figs.~\ref{fig:more_composition}(h, l), using $\mathbf{e}_{R}^{5} \cdot \mathbf{e}_{R}^{4} - \mathbf{e}_{R}^{6}$ and $\mathbf{e}_{R}^{4} \cdot \mathbf{e}_{R}^{5} - \mathbf{e}_{R}^{6}$ respectively, show nearly indistinguishable histograms, indicating that swapping the sequence of the two relations in OrthogonalE(2$\times$2) does not affect the outcome of the composition, suggesting it is commutative. Conversely, Figs.~\ref{fig:more_composition}(m, n), using $\mathbf{e}_{R}^{8} \cdot \mathbf{e}_{R}^{7} - \mathbf{e}_{R}^{9}$ and $\mathbf{e}_{R}^{7} \cdot \mathbf{e}_{R}^{8} - \mathbf{e}_{R}^{9}$, show that the former results in a trend closer to or equal to 0 more distinctly than the latter, implying that changing the sequence of relations affects the outcome, thereby demonstrating the non-commutative nature of relations in OrthogonalE(3$\times$3). In conclusion, even though OrthogonalE(2$\times$2) generally outperforms OrthogonalE(3$\times$3) in composition relation patterns, the comparative analysis reveals that OrthogonalE(3$\times$3) indeed possesses non-commutative composition properties, following the equation~\ref{equ:composition_proof} and \ref{equ:non_commutative}.

\subsection{Introduction of Relation Patterns}  \label{subsec:relation_pattern_explanations}

We can observe several relation patterns in KGs, including \textit{symmetry}, \textit{antisymmetry}, \textit{inversion}, and \textit{composition} (both commutative and non-commutative). Detailed examples have been shown in Fig.~\ref{fig:relation_pattern_examples}.

\paragraph{Symmetry and Antisymmetry}
Certain relations demonstrate \textit{symmetry}, indicating that the validity of a relation between entities $x$ and $y$ ($\left(r_{1}(x, y) \Rightarrow r_{1}(y, x)\right)$) (for instance, \textit{is married to}) is equally valid in the opposite direction (namely, from $y$ to $x$). Conversely, other relations are characterized by \textit{antisymmetry} ($\left(r_{1}(x, y) \Rightarrow \neg r_{1}(y, x)\right)$), signifying that if a relation is applicable between $x$ and $y$ (such as \textit{is father of}), it is inapplicable in the reverse direction (from $y$ to $x$).

\paragraph{Inversion}
Relations can also exhibit inversion ($\left(r_{1}(x, y) \Leftrightarrow r_{2}(y, x)\right)$), where reversing the direction of a relation effectively transforms it into another relation (for example, \textit{is child of} and \textit{is parent of}).

\paragraph{Composition}
The composition of relations ($\left(r_{1}(x, y) \cap r_{2}(y, z) \Rightarrow r_{3}(x, z)\right)$) signifies a crucial pattern wherein merging two or more relations facilitates the deduction of a novel relation. Such compositions might be commutative, where the sequence of relations is irrelevant, or non-commutative, where the sequence significantly influences the outcome. In scenarios where the order of relations is pivotal, as illustrated by the relationship where B is the mother of A's father and E is the father of A's mother, non-commutative composition ($\left(r_{1}(x, y) \cap r_{2}(y, z) \neq \right(r_{2}(x, y) \cap r_{1}(y, z)$) becomes essential. While commutative compositions would consider B and E as identical, non-commutative compositions recognize them as distinct.

%......................................table
\begin{table*}[t]
\resizebox{1\textwidth}{!}{\renewcommand{\arraystretch}{1}
    \centering
    \begin{tabular}{lcc}
    \clineB{1-3}{2}
     KGE Model & Description & MRR Accuracy \\
     \clineB{1-3}{2}
        MoCoSA\cite{he2023mocosa}       & Language Models & .696   \\ 
        SimKGC\cite{wang2022simkgc}     & Language Models & .671   \\
        LERP\cite{han2023logical}       & Additional Contextual Information (Logic Rules) & .622 \\
        C-LMKE\cite{wang2022language}   & Language Models & .598 \\
        KNN-KGE\cite{zhang2022reasoning}& Language Models & .579 \\
        HittER\cite{chen2020hitter}     & Transformer structure & .503 \\
        OrthogonalE($10 \times 10$)     & -               & .494 \\
    \clineB{1-3}{2}
    \end{tabular}
    }
    \caption{Other baseline models in WN18RR dataset.}
    \label{tab:SoTA_model_WN18RR}
\end{table*}
%......................................table

\subsection{Other baseline KGE model} \label{subsec:baseline_KGE}

In recent times, several significant performance methods have been developed, as detailed for WN18RR in Table~\ref{tab:SoTA_model_WN18RR}. Among these, MoCoSA\cite{he2023mocosa}, SimKGC\cite{wang2022simkgc}, C-LMKE\cite{wang2022language}, and KNN-KGE\cite{zhang2022reasoning} mainly utilize Language Models (LMs) to enrich dataset semantic information, thereby achieving superior outcomes. Conversely, LERP\cite{han2023logical} does not employ LMs but leverages additional contextual information (logic rules) beyond the dataset to fill information gaps in entities and relations, HittER \cite{chen2020hitter} integrates the transformer architecture into KGE, yet its lack of explainability remains unresolved. On the other hand, methods such as TransE\cite{bordes2013translating}, RotatE\cite{sun2019rotate}, and the OrthogonalE method introduced in this paper depend solely on the inherent data and information of the KGE dataset itself. These methods, based on specific mathematical rules and algorithms, do not incorporate any external information and thus do not operate as black-box approaches like LLMs. Consequently, these dataset-dependent methods remain highly valuable for KGE research.

\subsection{hyperparameter}
All the hyperparameter settings have been shown in Table~\ref{tab:hyperparameter}.

\end{document}